# Learning First-Order Definitions of Functions

**J. R. Quinlan**                                           QUINLAN@CS.SU.OZ.AU
*Basser Department of Computer Science*
*University of Sydney*
*Sydney 2006 Australia*

## Abstract

First-order learning involves finding a clause-form definition of a relation from examples of the relation and relevant background information. In this paper, a particular first-order learning system is modified to customize it for finding definitions of functional relations. This restriction leads to faster learning times and, in some cases, to definitions that have higher predictive accuracy. Other first-order learning systems might benefit from similar specialization.

## 1. Introduction

Empirical learning is the subfield of AI that develops algorithms for constructing theories from data. Most classification research in this area has used the *attribute-value* formalism, in which data are represented as vectors of values of a fixed set of attributes and are labelled with one of a small number of discrete classes. A learning system then develops a mapping from attribute values to classes that can be used to classify unseen data.

Despite the well-documented successes of algorithms developed for this paradigm (e.g., Michie, Spiegelhalter, and Taylor, 1994; Langley and Simon, 1995), there are potential applications of learning that do not fit within it. Data may concern objects or observations with arbitrarily complex structure that cannot be captured by the values of a predetermined set of attributes. Similarly, the propositional theory language employed by attribute-value learners may be inadequate to express patterns in such structured data. Instead, it may be necessary to describe learning input by *relations*, where a relation is just a set of tuples of constants, and to represent what is learned in a first-order language. Four examples of practical learning tasks of this kind are:

- *Speeding up logic programs* (Zelle and Mooney, 1993). The idea here is to learn a guard for each nondeterministic clause that inhibits its execution unless it will lead to a solution. Input to the learner consists of a Prolog program and one or more execution traces. In one example DOLPHIN, the system cited above, transformed a program from complexity $O(n!)$ to $O(n^2)$.

- *Learning search control heuristics* (Leckie and Zukerman, 1993). Formulation of preference criteria that improve efficiency in planning applications has a similar flavor. One task investigated here is the familiar 'blocks world' in which varying numbers of blocks must be rearranged by a robot manipulator. Each input to learning concerns a particular situation during search and includes a complete description of the current planning state and goals. As the amount of this information increases with the number of blocks and their inter-relationships, it cannot be encoded as a fixed set of values.

- *Recovering software specifications.* Cohen (1994) describes an application based on a software system consisting of over a million lines of C code. Part of the system implements virtual relations that compute projections and joins of the underlying base relations, and the goal is to reconstruct their definitions. Input to learning consists of queries, their responses, and traces showing the base relations accessed while answering the queries. The output is a logical description of the virtual relation; since this involves quantified variables, it lies beyond the scope of propositional attribute-value languages.

- *Learning properties of organic molecules* (Muggleton, King, and Sternberg, 1992; Srinivasan, Muggleton, Sternberg, and King, 1996). The approach to learning in these papers is based on representing the structure of the molecules themselves in addition to properties of molecules and molecule segments. The latter paper notes the discovery of a useful indicator of mutagenicity expressed in terms of this structure.

The development of learning methods based on this more powerful relational formalism is sometimes called *inductive logic programming* (Muggleton, 1992; Lavrač and Džeroski, 1994; De Raedt, 1996). Input typically consists of tuples that belong, or do not belong, to a *target* relation, together with relevant information expressed as a set of *background* relations. The learning task is then to formulate a definition of the target relation in terms of itself and the background relations.

This relational learning task is described in more detail in the following section. Several algorithms for relational learning have been developed recently, and Section 3 introduces one such system called FOIL (Quinlan, 1990). While FOIL can be used with relations of any kind, one particularly common use of relations is to represent functions. Changes to FOIL that in effect customize it for learning functional relations are outlined in Section 4. Several comparative studies, presented in Section 5, show that this specialization leads to much shorter learning times and, in some cases, to more accurate definitions. Related work on learning functional relations is discussed in Section 6, and the paper ends with some conclusions from this study and directions for further development.

## 2. Relational Learning

An $n$-ary relation $R_E$ consists of a set of $n$-tuples of ground terms (here constants). All constants in the $i$th position of the tuples belong to some type, where types may be differentiated or all constants may be taken to belong to a single universal type.

As an alternative to this extensional definition as a (possibly infinite) set, a relation can be specified intensionally via an $n$-argument predicate $R_I$ defined by a Prolog program. If

$$\langle c_1, c_2, ...c_n \rangle \in R_E \text{ if and only if } R_I(c_1, c_2, ..., c_n) \text{ is true}$$

for any constants $\{c_i\}$, then the intensional and extensional definitions are equivalent. For convenience, the subscripts of $R_E$ and $R_I$ will be omitted and $R$ will be used to denote either the set of tuples or the predicate.

Input to a relational learning task consists of extensional information about a target relation $R$ and extensional or intensional definitions of a collection of background relations. Examples of tuples known to belong to the target relation are provided and, in most cases, so are examples of tuples known not to belong to $R$. The goal is to learn a Prolog program for $R$ that covers all tuples known to belong to $R$ but no tuples known not to belong to $R$ or, in other words, a program that agrees with the extensional information provided about $R$.

Many relations of interest are infinite. An alternative to selecting examples that belong or do not belong to $R$ is to define a finite vocabulary $V$ and to specify relations with respect to this vocabulary. That is, $R$ is represented as the finite set of tuples, all constants in which belong to $V$. Since this specification of $R$ is complete over the vocabulary, the tuples that do not belong to $R$ can be inferred by the *closed world assumption* as the complement of the tuples in $R$.

A function $f(X_1, X_2, ..., X_k)$ of $k$ arguments can be represented by a $k+1$-ary relation $F(X_1, X_2, ..., X_k, X_{k+1})$ where, for each tuple in $F$, the value of the last argument is the result of applying $f$ to the first $k$ arguments. (Rouveirol (1994) proves that this *flattening* can be used to remove all non-constant function symbols from any first-order language.) Such *functional* relations have an additional property that for any constants $\{c_1, c_2, .., c_k\}$ there is exactly one value of $c_{k+1}$ such that $\langle c_1, c_2, ..., c_{k+1} \rangle$ belongs to $F$.

As an example, consider the three-argument predicate append(A,B,C) whose meaning is that the result of appending list A to list B is list C.[1] The corresponding relation append is infinite, but a restricted vocabulary can be defined as all flat lists containing only elements from {1,2,3} whose length is less than or equal to 3. There are 40 such lists

$$[\,], [1], [2], [3], [1,2], ...., [3,3,2], [3,3,3]$$

and 64,000 3-tuples of lists. With respect to this vocabulary, append consists of 142 of these 3-tuples, viz.:

$$\langle [\,],[\,],[\,]\rangle, \ \langle [\,],[1],[1]\rangle, \ ..., \ \langle [2],[1,3],[2,1,3]\rangle, \ ..., \ \langle [3,3,3],[\,],[3,3,3]\rangle.$$

There is also a background relation components, where components(A,B,C) means that list A has head B and tail C. The goal is then to learn an intensional definition of append given the background relation components. A suitable result might be expressed as

```
append([ ],A,A).
append(A,B,C) :- components(A,D,E), append(E,B,F), components(C,D,F).
```

which is recognizable as a Prolog definition of append.

---

1. In Prolog, append can be invoked with any combination of its arguments bound so as to find possible values for the unbound arguments. Here and in Section 5.1, however, append is treated as a function from the first two arguments to the third.

```
Initialization:
    definition := null program
    remaining := all tuples belonging to target relation R

While remaining is not empty
    /* Grow a new clause */
    clause := R(A, B, ...) :-
        While clause covers tuples known not to belong to R
            /* Specialize clause */
            Find appropriate literal(s) L
            Add L to body of clause
    Remove from remaining tuples in R covered by clause
    Add clause to definition
```

*Figure 1:* Outline of FOIL

## 3. Description of FOIL

In common with many first-order learning systems, FOIL requires that the background relations are also defined extensionally by sets of tuples of constants.[2] Although the intensional definition is learned from a particular set of examples, it is intended to be executable as a Prolog program in which the background relations may also be specified intensionally by definitions rather than by sets of ground tuples. For instance, although the append definition above might have been learned from particular examples of lists, it will correctly append arbitrary lists, provided that components is specified by a suitable clausal definition. (The applicability of learned definitions to unseen examples cannot be guaranteed, however; Bell and Weber (1993) call this the *open domain assumption*.)

The language in which FOIL expresses theories is a restricted form of Prolog that omits cuts, fail, disjunctive goals, and functions other than constants, but allows negated literals *not(L(...))*. This is essentially the *Datalog* language specified by Ullman (1988), except that there is no requirement that all variables in a negated literal appear also in the head or in another unnegated literal; FOIL interprets *not* using *negation as failure* (Bratko, 1990).

### 3.1 Broad-brush overview

As outlined in Figure 1, FOIL uses the separate-and-conquer method, iteratively learning a clause and removing the tuples in the target relation $R$ covered by the clause until none remain. A clause is grown by repeated specialization, starting with the most general clause head and adding literals to the body until the clause does not cover any tuples known not to belong to $R$.

---

2. Prominent exceptions include FOCL (Pazzani and Kibler, 1992), FILP (Bergadano and Gunetti, 1993), and FOIDL (Mooney and Califf, 1995), that allow background relations to be defined extensionally, and Progol (Muggleton, 1995), in which information about all relations can be in non-ground form.

Literals that can appear in the body of a clause are restricted by the requirement that programs be function-free, other than for constants appearing in equalities. The possible literal forms that FOIL considers are:

- $Q(X_1, X_2, ..., X_k)$ and $not(Q(X_1, X_2, ..., X_k))$, where $Q$ is a relation and the $X_i$'s denote *known* variables that have been bound earlier in the clause or *new* variables. At least one variable must have been bound earlier in the partial clause, either by the head or a literal in the body.

- $X_i{=}X_j$ and $X_i{\neq}X_j$, for known variables $X_i$ and $X_j$ of the same type.

- $X_i{=}c$ and $X_i{\neq}c$, where $X_i$ is a known variable and $c$ is a constant of the appropriate type. Only constants that have been designated as suitable to appear in a definition are considered – a reasonable definition for `append` might reference the null list [ ] but not an arbitrary list such as [1,2].

- $X_i \leq X_j$, $X_i > X_j$, $X_i \leq t$, and $X_i > t$, where $X_i$ and $X_j$ are known variables with numeric values and $t$ is a threshold chosen by FOIL.

If the learned definition must be pure Prolog, negated literal forms $not(Q(...))$ and $X_i{\neq}...$ can be excluded by an option.

Clause construction is guided by different possible *bindings* of the variables in the partial clause that satisfy the clause body. If the clause contains $k$ variables, a binding is a $k$-tuple of constants that specifies a value for all variables in sequence. Each possible binding is labelled $\oplus$ or $\ominus$ according to whether the tuple of values for the variables in the clause head does or does not belong in the target relation.

As an illustration, consider the tiny task of constructing a definition of plus(A,B,C), meaning A+B = C, using the background relation dec(A,B), denoting B = A−1. The vocabulary is restricted to just the integers 0, 1, and 2, so that plus consists of the tuples

$$\langle 0,0,0 \rangle, \ \langle 1,0,1 \rangle, \ \langle 2,0,2 \rangle, \ \langle 0,1,1 \rangle, \ \langle 1,1,2 \rangle, \ \langle 0,2,2 \rangle$$

and dec contains only $\langle 1,0 \rangle$ and $\langle 2,1 \rangle$.

The initial clause consists of the head

plus(A,B,C) :-

in which each variable is unique. The labelled bindings corresponding to this initial partial clause are just the tuples that belong, or do not belong, to the target relation, i.e.:

| | | | | | |
|---|---|---|---|---|---|
| $\langle 0,0,0 \rangle \oplus$ | $\langle 1,0,1 \rangle \oplus$ | $\langle 2,0,2 \rangle \oplus$ | $\langle 0,1,1 \rangle \oplus$ | $\langle 1,1,2 \rangle \oplus$ | $\langle 0,2,2 \rangle \oplus$ |
| $\langle 0,0,1 \rangle \ominus$ | $\langle 0,0,2 \rangle \ominus$ | $\langle 0,1,0 \rangle \ominus$ | $\langle 0,1,2 \rangle \ominus$ | $\langle 0,2,0 \rangle \ominus$ | $\langle 0,2,1 \rangle \ominus$ |
| $\langle 1,0,0 \rangle \ominus$ | $\langle 1,0,2 \rangle \ominus$ | $\langle 1,1,0 \rangle \ominus$ | $\langle 1,1,1 \rangle \ominus$ | $\langle 1,2,0 \rangle \ominus$ | $\langle 1,2,1 \rangle \ominus$ |
| $\langle 1,2,2 \rangle \ominus$ | $\langle 2,0,0 \rangle \ominus$ | $\langle 2,0,1 \rangle \ominus$ | $\langle 2,1,0 \rangle \ominus$ | $\langle 2,1,1 \rangle \ominus$ | $\langle 2,1,2 \rangle \ominus$ |
| $\langle 2,2,0 \rangle \ominus$ | $\langle 2,2,1 \rangle \ominus$ | $\langle 2,2,2 \rangle \ominus$ | | | . |

FOIL repeatedly tries to construct a clause that covers some tuples in the target relation $R$ but no tuples that are definitely not in $R$. This can be restated as finding a clause that has some $\oplus$ bindings but no $\ominus$ bindings, so one reason for adding a literal to the clause is to move in this direction by increasing the relative proportion of $\oplus$ bindings. Such *gainful* literals are evaluated using an information-based heuristic. Let the number of $\oplus$ and $\ominus$ bindings of a partial clause be $n^{\oplus}$ and $n^{\ominus}$ respectively. The average information provided by the discovery that one of the bindings has label $\oplus$ is

$$I(n^{\oplus}, n^{\ominus}) = - \log_2 \left( \frac{n^{\oplus}}{n^{\oplus} + n^{\ominus}} \right) \text{ bits.}$$

If a literal $L$ is added, some of these bindings may be excluded and each of the rest will give rise to one or more bindings for the new partial clause. Suppose that $k$ of the $n^{\oplus}$ bindings are not excluded by $L$, and that the numbers of bindings of the new partial clause are $m^{\oplus}$ and $m^{\ominus}$ respectively. If $L$ is chosen so as to increase the proportion of $\oplus$ bindings, the total information gained by adding $L$ is then

$$k \times (I(n^{\oplus}, n^{\ominus}) - I(m^{\oplus}, m^{\ominus})) \text{ bits.}$$

Consider the result of specializing the above clause by the addition of the literal A=0. All but nine of the bindings are eliminated because the corresponding values of the variables do not satisfy the new partial clause. The bindings are reduced to

| ⟨0,0,0⟩ $\oplus$ | ⟨0,1,1⟩ $\oplus$ | ⟨0,2,2⟩ $\oplus$ | | | |
|---|---|---|---|---|---|
| ⟨0,0,1⟩ $\ominus$ | ⟨0,0,2⟩ $\ominus$ | ⟨0,1,0⟩ $\ominus$ | ⟨0,1,2⟩ $\ominus$ | ⟨0,2,0⟩ $\ominus$ | ⟨0,2,1⟩ $\ominus$ |

in which the proportion of $\oplus$ bindings has increased from 6/27 to 3/9. The information gained by adding this literal is therefore $3 \times (I(6, 21) - I(3, 6))$ or about 2 bits. Adding the further literal B=C excludes all the $\ominus$ bindings, giving a complete first clause

plus(A,B,C) :- A=0, B=C.

or, as it would be more commonly written,

plus(0,B,B).

This clause covers three tuples of plus which are then removed from the set of tuples to be covered by subsequent clauses. At the commencement of the search for the second clause, the head is again plus(A,B,C) and the bindings are now

| ⟨1,0,1⟩ $\oplus$ | ⟨2,0,2⟩ $\oplus$ | ⟨1,1,2⟩ $\oplus$ | | | |
|---|---|---|---|---|---|
| ⟨0,0,1⟩ $\ominus$ | ⟨0,0,2⟩ $\ominus$ | ⟨0,1,0⟩ $\ominus$ | ⟨0,1,2⟩ $\ominus$ | ⟨0,2,0⟩ $\ominus$ | ⟨0,2,1⟩ $\ominus$ |
| ⟨1,0,0⟩ $\ominus$ | ⟨1,0,2⟩ $\ominus$ | ⟨1,1,0⟩ $\ominus$ | ⟨1,1,1⟩ $\ominus$ | ⟨1,2,0⟩ $\ominus$ | ⟨1,2,1⟩ $\ominus$ |
| ⟨1,2,2⟩ $\ominus$ | ⟨2,0,0⟩ $\ominus$ | ⟨2,0,1⟩ $\ominus$ | ⟨2,1,0⟩ $\ominus$ | ⟨2,1,1⟩ $\ominus$ | ⟨2,1,2⟩ $\ominus$ |
| ⟨2,2,0⟩ $\ominus$ | ⟨2,2,1⟩ $\ominus$ | ⟨2,2,2⟩ $\ominus$ | | | |

The above literals were added to the body of the first clause because they gain information. A quite different justification for adding a literal is to introduce new variables that may be needed in the final clause. *Determinate* literals are based on an idea introduced by Golem (Muggleton and Feng, 1992). A determinate literal is one that introduces new variables so that the new partial clause has exactly one binding for each $\oplus$ binding in the current clause, and at most one binding for each $\ominus$ binding. Determinate literals are useful because they introduce new variables, but neither reduce the potential coverage of the clause nor increase the number of bindings.

Now that the $\oplus$ bindings do not include any with A=0, the literal dec(A,D) is determinate because, for each value of A, there is one value of D that satisfies the literal. Similarly, since the $\oplus$ bindings contain none with C=0, the literal dec(C,E) is also determinate.

In Figure 1, the literals $L$ added by FOIL at each step are

- the literal with greatest gain, if this gain is near the maximum possible (namely $n^\oplus \times I(n^\oplus, n^\ominus)$) ; otherwise

- all determinate literals found; otherwise

- the literal with highest positive gain; otherwise

- the first literal investigated that introduces a new variable.

At the start of the second clause, no literal has near-maximum gain and so all determinate literals are added to the clause body. The partial clause

plus(A,B,C) :- dec(A,D), dec(C,E),

has five variables and the bindings that satisfy it are

| | | |
|---|---|---|
| $\langle 1,0,1,0,0 \rangle$ $\oplus$ | $\langle 2,0,2,1,1 \rangle$ $\oplus$ | $\langle 1,1,2,0,1 \rangle$ $\oplus$ |
| $\langle 1,0,2,0,1 \rangle$ $\ominus$ | $\langle 1,1,1,0,0 \rangle$ $\ominus$ | $\langle 2,0,1,1,0 \rangle$ $\ominus$ | $\langle 2,1,1,1,0 \rangle$ $\ominus$ | $\langle 2,1,2,1,1 \rangle$ $\ominus$ |

The literal plus(B,D,E), which uses these newly-introduced variables, is now satisfied by all three $\oplus$ bindings but none of the $\ominus$ bindings, giving a complete second clause

plus(A,B,C) :- dec(A,D), dec(C,E), plus(B,D,E).

All tuples in plus are covered by one or other of these clauses, so they constitute a complete intensional definition of the target relation.

## 3.2 Details omitted

FOIL is a good deal more complex than this overview would suggest. Since they are not important for this paper, matters such as the following are not discussed here, but are covered in (Quinlan and Cameron-Jones, 1993; 1995):

- *Recursive soundness.* If the goal is to be able to execute the learned definitions as ordinary Prolog programs, it is important that they terminate. FOIL has an elaborate mechanism to ensure that any recursive literal (such as plus(B,D,E) above) that is added to a clause body will not cause problems in this respect, at least for ground queries.

- *Pruning.* In practical applications with numerous background relations, the number of possible literals $L$ that could be added at each step grows exponentially with the number of variables in the partial clause. FOIL employs some further heuristics to limit this space, such as Golem's bound on the *depth* of a variable (Muggleton and Feng, 1992). More importantly, some regions of the literal space can be pruned without examination because they can be shown to contain neither determinate literals, nor literals with higher gain than the best gainful literal found so far.

- *More complete search.* As presented above, FOIL is a straightforward greedy hill-climbing algorithm. In fact, because FOIL can sometimes reach an impasse in its search for a clause, it contains a limited non-chronological backtracking facility to allow it to recover from such situations.

- *Simplifying definitions.* The addition to the partial clause of *all* determinate literals found may seem excessive. However, as a clause is completed, FOIL examines each literal in the clause body to see whether it could be discarded without causing the simpler clause to match tuples not in the target relation $R$. Similarly, when the definition is complete, each clause is checked to see whether it could be omitted without leaving any tuples in $R$ uncovered. There are also heuristics that aim to make clauses more understandable by substituting simpler literals (such as variable equalities) for literals based on more complex relations.

- *Recognizing boundaries of closed worlds.* Some literals that appear to discriminate $\oplus$ from $\ominus$ bindings do so only as a consequence of boundary effects attributable to a limited vocabulary.[3] When a definition including such literals is executed with larger vocabularies, the open domain assumption mentioned above may be violated. FOIL contains an optional mechanism for describing when literals might be satisfied by bindings outside the closed world, allowing some literals with unpredictable behavior to be excluded.

Quinlan (1990) and Quinlan and Cameron-Jones (1995) summarize several applications successfully addressed by FOIL, some of which are also discussed in Section 5.

## 4. Learning Functional Relations

The learning approach used by FOIL makes no assumptions about the form of the target relation $R$. However, as with append and plus above, the relation is often used to represent a function – in any tuple of constants that satisfies $R$, the last constant is uniquely determined by the others. Bergadano and Gunetti (1993) show that this property can be exploited to make the learning task more tractable.

### 4.1 Functional relations and FOIL

Although FOIL can learn definitions for functional relations, it is handicapped in two ways:

- *Ground queries:* FOIL's approach to recursive soundness assumes that only ground queries will be made of the learned definition. That is, a definition of $R(X_1, X_2, ..., X_n)$ will be used to provide true-false answers to queries of the form $R(c_1, c_2, ..., c_n)$? where the $c_i$'s are constants. If $R$ is a functional relation, however, a more sensible query would seem to be $R(c_1, c_2, ..., c_{n-1}, X)$? to determine the value of the function for specified ground arguments. In the case of plus, for instance, we would not expect to ask plus(1,1,2)? ("is 1+1=2?"), but rather plus(1,1,X)? ("what is 1+1?"). $R(c_1, c_2, ..., c_{n-1}, X)$? will be called the *standard query* for functional relations.

---

3. An example of this arises in Section 5.1.

- *Negative examples:* FOIL needs both tuples that belong to the target relation and at least some that do not. In common with other ILP systems such as Golem (Muggleton and Feng, 1992), the latter are used to detect when a partial clause is still too general. These can be specified to FOIL directly or, more commonly, are derived under the closed world assumption that, with respect to the vocabulary, all tuples in $R$ have been given. This second mechanism can often lead to very large collections of tuples not in $R$; there were nearly 64,000 of them in the append illustration earlier. Every tuple not belonging to $R$ results in a binding at the start of each clause, so there can be uncomfortably many bindings that must be maintained and tested at each stage of clause development.[4] However, functional relations do not need explicit counter-examples, even when the set of tuples belonging to $R$ is not complete with respect to some vocabulary – knowing that $\langle c_1, c_2, ..., c_n \rangle$ belongs to $R$ implies that there is no other constant $c'_n$ such that $\langle c_1, c_2, ..., c'_n \rangle$ is in $R$.

These problematic aspects of FOIL vis à vis functional relations suggest modifications to address them. The alterations lead to a new system, FFOIL, that is still close in spirit to its progenitor.

## 4.2 Description of FFOIL

Since the last argument of a functional relation has a special role, it will be referred to as the *output argument* of the relation. Similarly, the variable corresponding to this argument in the head of a clause is called the *output variable*.

The most fundamental change in FFOIL concerns the bindings of partial clauses and the way that they are labelled. A new constant $\square$ is introduced to indicate an undetermined value of the output variable in a binding. Bindings will be labelled according to the value of the output variable, namely $\oplus$ if this value is correct (given the value of the earlier constants), $\ominus$ if the value is incorrect, and $\odot$ if the value is undetermined.

The outline of FFOIL (Figure 2) is very similar to Figure 1, the only differences being those highlighted. At the start of each clause there is one binding for every remaining tuple in the target relation. The output variable has the value $\square$ in these bindings and this value is changed only when some subsequent literal assigns a value to the variable. In the small plus example of Section 3.1, the initial bindings for the first clause are

$$\langle 0,0,\square \rangle \; \odot \quad \langle 1,0,\square \rangle \; \odot \quad \langle 2,0,\square \rangle \; \odot \quad \langle 0,1,\square \rangle \; \odot \quad \langle 1,1,\square \rangle \; \odot \quad \langle 0,2,\square \rangle \; \odot$$

Like its ancestor, FFOIL also assesses potential literals for adding to the clause body as gainful or determinate, although both concepts must be adjusted to accommodate the new label $\odot$. Suppose that there are $r$ distinct constants in the range of the target function.

---

4. For this reason, FOIL includes an option to sample the $\ominus$ bindings instead of using all of them.

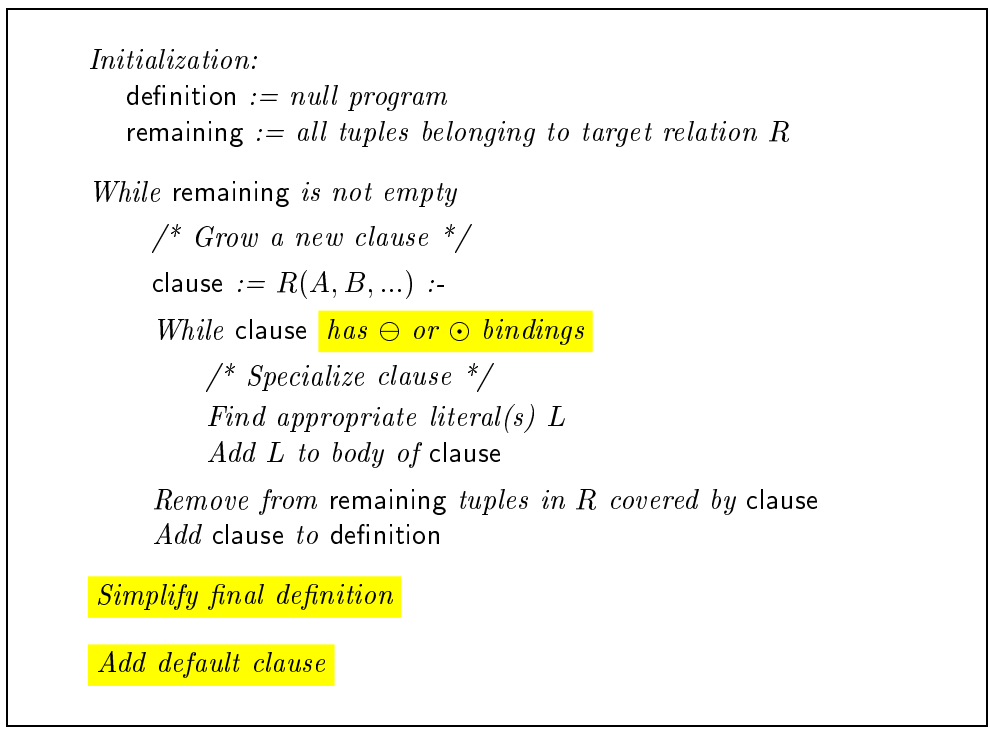

*Figure 2:* Outline of FFOIL

Each $\odot$ binding can be converted to a $\oplus$ binding only by changing $\square$ to the correct value of the function, and to a $\ominus$ binding by changing $\square$ to any of the $r-1$ incorrect values. In computing information gain, FFOIL thus counts each $\odot$ binding as 1 $\oplus$ binding and $r-1$ $\ominus$ bindings. A determinate literal is now one that introduces one or more variables so that, in the new partial clause, there is exactly one binding for each current $\oplus$ or $\odot$ binding and at most one binding for each current $\ominus$ binding. FFOIL uses the same preference criterion for adding literals $L$: a literal with near-maximum gain, then all determinate literals, then the most gainful literal, and finally a non-determinate literal that introduces a new variable.

The first literal chosen by FOIL in Section 3.1 was A=0 since this increases the concentration of $\oplus$ bindings from 9 in 64 to 3 in 9 (with a corresponding information gain). From FFOIL's perspective, however, this literal simply reduces six $\odot$ bindings to three and so gives no gain; as the range of plus is the set {0,1,2}, $r=3$, and so the putative concentration of $\oplus$ bindings would alter from 6 in 18 to 3 in 9. The literal A=C, on the other hand, causes the value of the output variable to be determined and results in the bindings

$$\frac{\langle 0,0,0\rangle \oplus \quad \langle 1,0,1\rangle \oplus \quad \langle 2,0,2\rangle \oplus}{\langle 0,1,0\rangle \ominus \quad \langle 1,1,1\rangle \ominus \quad \langle 0,2,0\rangle \ominus} .$$

This corresponds to an increase in concentration of $\oplus$ bindings from a notional 6 in 18 to 3 in 6, with an information gain of about 2 bits. Once this literal has been added to the clause body, FFOIL finds that a further literal B=0 eliminates all $\ominus$ bindings, giving the

complete clause

$$\text{plus}(\text{A},\text{B},\text{C}) :\!- \text{A}=\text{C}, \text{B}=0.$$

The remaining tuples of plus give the bindings

$$\langle 0,1,\square \rangle \odot \quad \langle 1,1,\square \rangle \odot \quad \langle 0,2,\square \rangle \odot$$

at the start of the second clause. The literals dec(B,D) and dec(E,A) are both determinate and, when they are added to the clause, the bindings become

$$\langle 0,1,\square,0,1 \rangle \odot \quad \langle 1,1,\square,0,2 \rangle \odot \quad \langle 0,2,\square,1,1 \rangle \odot$$

in which the output variable is still undetermined. If the partial clause is further specialized by adding the literal plus(E,D,C), the new bindings

$$\langle 0,1,1,0,1 \rangle \oplus \quad \langle 1,1,2,0,2 \rangle \oplus \quad \langle 0,2,2,1,1 \rangle \oplus$$

give the correct value for C in each case. Since there are no $\odot$ or $\ominus$ bindings, this clause is also complete.

One important consequence of the new way that bindings are initialized at the start of each clause is easily overlooked. With FOIL, there is one $\ominus$ binding for each tuple that does not belong in $R$; since each clause excludes all $\ominus$ bindings, it discriminates some tuples in $R$ from all tuples not in $R$. This is the reason that the learned clauses can be regarded as a set and can be executed in any order without changing the set of answers to a query. In FFOIL, however, the initial bindings concern only the remaining tuples in $R$, so a learned clause depends on the context established by earlier clauses. For example, suppose a target relation S and a background relation T are defined as

$$\text{S} = \{\langle v,1 \rangle, \langle w,1 \rangle, \langle x,1 \rangle, \langle y,0 \rangle, \langle z,0 \rangle\}$$
$$\text{T} = \{\langle v \rangle, \langle w \rangle, \langle x \rangle\} \ .$$

The first clause learned by FFOIL might be

$$\text{S}(\text{A},1) :\!- \text{T}(\text{A}).$$

and the remaining bindings $\{\langle y,0 \rangle, \langle z,0 \rangle\}$ could then be covered by the clause

$$\text{S}(\text{A},0).$$

The latter clause is clearly correct only for standard queries that are not covered by the first clause. As this example illustrates, the learned clauses must be interpreted in the order in which they were learned, and each clause must be ended with a cut '!' to protect later clauses from giving possibly incorrect answers to a query. Since the target relation $R$ is functional, so that there is only one correct response to a standard query as defined above, this use of cuts is safe in that it cannot rule out a correct answer.

Both FOIL and FFOIL tend to give up too easily when learning definitions to explain noisy data. This can result in over-specialized clauses that cover the target relation only partially. On tasks for which the definition learned by FFOIL is incomplete, a final *global simplification* phase is invoked. Clauses in the definition are generalized by removing literals so long as the total number of errors on the target relation does not increase. In this way, the accuracy of individual clauses is balanced against the accuracy of the definition as a whole; simplifying a clause by removing a literal may increase the number of errors made by the clause, but this can be offset by a reduction in the number of uncovered bindings and a consequently lower global error rate. When all clauses have been simplified as much as possible, entire clauses that contribute nothing to the accuracy of the definition are removed.

In the final step of Figure 2, the target relation is assumed to represent a total function, with the consequence that a response must always be returned for a standard query. As a safeguard, FFOIL adds a default clause

$$R(X_1, X_2, ..., X_{n-1}, c).$$

where $c$ is the most common value of the function.[5] The most common value of the output argument of plus is 2, so the complete definition for this example, in normal Prolog notation, becomes

```
plus(A,0,A) :- !.
plus(A,B,C) :- dec(B,D), dec(E,A), plus(E,D,C), !.
plus(A,B,2).
```

## 4.3 Advantages and disadvantages of FFOIL

Although the definitions for plus in Sections 3.1 and 4.2 are superficially similar, there are considerable differences in the learning processes by which they were constructed and in their operational characteristics when used.

- FFOIL generally needs to maintain fewer bindings and so learns more quickly. Whereas FOIL keeps up to 27 bindings while learning a definition of plus, FFOIL never uses more than 6.

- The output variable is guaranteed to be bound in every clause learned by FFOIL. This is not necessarily the case with FOIL, since there is no requirement that every variable appearing in the head must also appear in the clause body.

- Definitions found by FFOIL often execute more efficiently than their FOIL counterparts. Firstly, FFOIL definitions, through the use of cuts, exploit the fact that there cannot be more than one correct answer to a standard query. Secondly, clause bodies constructed by FFOIL tend not to use the output variable until it has been bound, so there is less backtracking during evaluation. As an illustration, the FOIL definition of Section 3.1 evaluates 81 goals in answering the query plus(1,1,X)?, many more than the six evaluations needed by the FFOIL definition for the same query.

There are also entries on the other side of the ledger:

---

5. No default clause is added if each value of the function occurs only once.

| Task | Bkgd Relns | Length 3 | | | | Length 4 | | | |
|---|---|---|---|---|---|---|---|---|---|
| | | Bindings | | Time | | Bindings | | Time | |
| | | $\oplus$ | $\ominus$ | FOIL | FFOIL | $\oplus$ | $\ominus$ | FOIL | FFOIL |
| append | 2 | 142 | 63,858 | 3.0 | 0.5 | 1593 | 396,502 | 22.4 | 10.9 |
| last element | 3 | 39 | 81 | 0.0 | 0.0 | 340 | 1024 | 0.5 | 0.3 |
| reverse | 10 | 40 | 1560 | 2.6 | 0.3 | 341 | 115,940 | 195.9 | 9.0 |
| left shift | 12 | 39 | 1561 | 0.5 | 0.3 | 340 | 115,940 | 26.6 | 6.8 |
| translate | 14 | 40 | 3120 | 817.9 | 1.1 | 341 | 115,940 | 495.9 | 28.0 |

*Table 1:* Results on tasks from (Bratko, 1990).

- FOIL is applicable to more learning tasks that FFOIL, which is limited to learning definitions of functional relations.

- The implementation of FFOIL is more complex than that of FOIL. For example, many of the heuristics for pruning the literal search space and for checking recursive soundness require special cases for the constant □ and for ⊙ bindings.

## 5. Empirical Trials

In this section the performance of FFOIL on a variety of learning tasks is summarized and compared with that of FOIL (release 6.4). Since the systems are similar in most respects, this comparison highlights the consequences of restricting the target relation to a function. Times are for a DEC AXP 3000/900 workstation. The learned definitions from the first three subsections may be found in the Appendix.

### 5.1 Small list manipulation programs

Quinlan and Cameron-Jones (1993) report the results of applying FOIL to 16 tasks taken from Bratko's (1990) well-known Prolog text. The list-processing examples and exercises of Chapter 3 are attempted in sequence, where the background information for each task includes all previously-encountered relations (even though most of them are irrelevant to the task at hand). Two different vocabularies are used: all 40 lists of length up to 3 on three elements and all 341 lists of length up to 4 on four elements.

Table 1 describes the five functional relations in this set and presents the performance of FOIL and FFOIL on them. All the learned definitions are correct for arbitrary lists, with one exception – FOIL's definition of reverse learned from the larger vocabulary includes the clause

$$\text{reverse(A,A) :- append(A,A,C), del(D,E,C).}$$

that exploits the bounded length of lists.[6] The times reveal a considerable advantage to FFOIL in all tasks except the second. In fact, for the first and last task with the larger vocabulary, these times understate FFOIL's advantage. The total number of bindings for append is $341^3$, or about 40 million, so a FOIL option was used to sample only 1% of the $\ominus$ bindings to prevent FOIL exceeding available memory. Had it been possible to run FOIL with all bindings, the time required to learn the definition would have been considerably longer. Similarly, FOIL exhausted available memory on the translation task when all 232,221 possible bindings were used, so the above results were obtained using a sample of 50% of the $\ominus$ bindings.

---

6. If C is twice the length of A and E is one element longer than C while still having length $\leq 4$, then the length of A must be 0 or 1. In that case A is its own reverse.

| Task | FOIL | FFOIL |
|------|------|-------|
| quicksort | 4.7 | 2.2 |
| bubblesort | 7.3 | 0.4 |

*Table 2:* Times (sec) for learning to sort.

| | Time (secs) | | | | Ratio to [3,3] | | |
|--------|------|------|--------|--------|------|------|---------|
| | [3,3] | [3,4] | [4,4] | [4,5] | [3,4] | [4,4] | [4,5] |
| FFOIL | 0.7 | 1.5 | 4.5 | 15.0 | 2.1 | 6.4 | 21.4 |
| FOIL | 0.8 | 4.3 | 11.9 | 146.3 | 5.4 | 14.9 | 182.9 |
| Golem | 4.8 | 14.6 | 59.6 | >395 | 3.0 | 12.4 | >82.3 |
| Progol | 43.0 | 447.9 | 5271.9 | >76575 | 10.4 | 122.6 | >1780.8 |

*Table 3:* Comparative times for quicksort task.

## 5.2 Learning quicksort and bubblesort

These tasks concern learning how to sort lists from examples of sorted lists. In the first, the target relation qsort(A,B) means that B is the sorted form of A. Three background relations are provided: components and append as before, and partition(A,B,C,D), meaning that partitioning list B on value A gives the list C of elements less than A and list D of elements greater than A. In the second task, the only background relations for learning bsort(A,B) are components and lt(A,B), meaning A<B. The vocabulary used for both tasks is all lists of length up to 4 with non-repeated elements drawn from {1,2,3,4}. There are thus 65 $\oplus$ and 4160 $\ominus$ bindings for each task.

Both FOIL and FFOIL learn the "standard" definition of quicksort. Times shown in Table 2 are comparable, mainly because FFOIL learns a superfluous over-specialized clause that is later discarded in favor of the more general recursive clause. The outcome for bubblesort is quite different – FFOIL learns twenty times faster than FOIL but its definition is more verbose.

The quicksort task provides an opportunity to compare FFOIL with two other well-known relational learning systems. Like FFOIL and FOIL, both Golem (Muggleton and

| Task | FOIL | FFOIL |
|------|------|-------|
| Ackermann's function | 12.3 | 0.2 |
| greatest common divisor | 237.5 | 1.2 |

*Table 4:* Times (sec) for arithmetic functions.

Feng, 1992) and Progol (release 4.1) (Muggleton, 1995) are implemented in C, so that timing comparisons are meaningful. Furthermore, both systems include quicksort among their demonstration learning tasks, so it is reasonable to assume that the parameters that control these systems have been set to appropriate values.

The four learning systems are evaluated using four sets of training examples, obtained by varying the maximum length $S$ of the lists and the size $A$ of the alphabet of non-repeating elements that can appear in the lists, as in (Quinlan, 1991). Denoting each set by a pair $[S,A]$, the four datasets are [3,3], [3,4], [4,4], and [4,5]. The total numbers of possible bindings for these tasks, 256, 1681, 4225, and 42,436 respectively, span two orders of magnitude. Table 3 summarizes the execution times[7] required by the systems on these datasets. Neither Golem nor Progol completed the last task; Golem exhausted the available swap space of 60Mb, and Progol was terminated after using nearly a day of cpu time. The table also shows the ratio of the execution time of the latter three to the simplest dataset [3,3]. The growth in FFOIL's execution time is far slower than that of the other systems, primarily because FFOIL needs only the $\oplus$ tuples while the others use both $\oplus$ and $\ominus$ tuples. Golem's execution time seems to grow slightly slower than FOIL's, while Progol's growth rate is much higher.

### 5.3 Arithmetic functions

The systems have been also used to learn definitions of complex functions from arithmetic. Ackermann's function

$$f(m,n) = \begin{cases} n+1 & \text{if } m = 0 \\ f(m-1,1) & \text{if } n = 0 \\ f(m-1, f(m,n-1)) & \text{otherwise} \end{cases}$$

provides a testing example for recursion control; the background relation succ(A,B) represents B=A+1. Finding the greatest common divisor of two numbers is another interesting task; the background relation is plus. For these tasks the vocabulary consists of the integers 0 to 20 and 1 to 20 respectively, giving 51 tuples in Ackermann(A,B,C) such that A, B and C are all less than or equal to 20, and 400 tuples in gcd(A,B,C).

As shown in Table 4, FFOIL is about 60 times faster than FOIL when learning a definition for Ackermann and about 200 times faster for gcd. This is due solely to FFOIL's smaller numbers of bindings. In gcd, for example, FOIL starts with $20^3$ or 8,000 bindings whereas FFOIL never uses more than 400 bindings.

---

7. Difficulties were experienced running Golem on an AXP 3000/900, so all times in this table are for a DECstation 5000/260.

Both FOIL and FFOIL learn exactly the same program for Ackermann's function that mirrors the definition above. In the case of gcd, however, the definitions highlight the potential simplification achievable with ordered clauses. The definition found by FOIL is

    gcd(A,A,A).
    gcd(A,B,C) :- plus(B,D,A), gcd(B,A,C).
    gcd(A,B,C) :- plus(A,D,B), gcd(A,D,C).

while that learned by FFOIL (omitting the default clause) is

    gcd(A,A,A) :- !.
    gcd(A,B,C) :- plus(A,D,B), gcd(A,D,C), !.
    gcd(A,B,C) :- gcd(B,A,C), !.

The last clause exploits the fact that all cases in which A is less than or equal to B have been filtered out by the first two clauses.

## 5.4 Finding the past tense of English verbs

The previous examples have all concerned tasks for which a compact, correct definition is known to exist. This application, learning how to change an English verb in phonetic notation from present to past tense, has more of a real-world flavor in that any totally correct definition would be extremely complex. A considerable literature has built up around this task, starting in the connectionist community, moving to symbolic learning through the work of Ling (1994), then to relational learning (Quinlan, 1994; Mooney and Califf, 1995).

Quinlan (1994) proposes representing this task as a relation past(A,B,C), interpreted as the past tense of verb A is formed by stripping off the ending B and then adding string C. The single background relation split(A,B,C) shows all ways in which word A can be split into two non-empty substrings B and C. Following the experiment reported in (Ling, 1994), a corpus of 1391 verbs is used to generate ten randomly-selected learning tasks, each containing 500 verbs from which a definition is learned and 500 different verbs used to test the definition. A Prolog interpreter is used to evaluate the definitions learned by FOIL, each unseen word w being mapped to a test query past(w,X,Y)?. The result of this query is judged correct only when both X and Y are bound to the proper strings. If there are multiple responses to the query, only the first is used – this disadvantages FOIL somewhat, since the system does not attempt to reorder learned clauses for maximum accuracy on single-response queries. The average accuracy of the definitions found by FOIL is 83.7%.

To apply FFOIL to this task, the relation past(A,B,C) must be factored into two functional relations delete(A,B) and add(A,C) since FFOIL can currently learn only functions with a single output variable. The same training and test sets of verbs are used, each giving rise to two separate learning tasks, and a test is judged correct only when both delete and add give the correct results for the unseen verb. The definitions learned by FFOIL have a higher average accuracy of 88.9%; on the ten trials, FFOIL outperforms FOIL on nine and is inferior on one, so the difference is significant at about the 1% level using a one-tailed sign test. The average time required by FFOIL to learn a pair of definitions, approximately 7.5 minutes, is somewhat less than the time taken by FOIL to learn a single definition.

| Object | Edges | Correct | | | | | Time (sec) | |
|--------|-------|------|-------|-------|-------|-------|------|-------|
| | | FOIL | FFOIL | mFOIL | Golem | FORS | FOIL | FFOIL |
| A | 54 | 16 | 21 | 22 | 17 | 22 | 2.5 | 9.1 |
| B | 42 | 9 | 15 | 12 | 9 | 12 | 1.7 | 11.0 |
| C | 28 | 8 | 11 | 9 | 5 | 8 | 3.3 | 9.7 |
| D | 57 | 10 | 22 | 6 | 11 | 16 | 2.4 | 11.1 |
| E | 96 | 16 | 54 | 10 | 10 | 29 | 4.7 | 5.9 |
| Total | 277 | 59 | 123 | 59 | 52 | 87 | 14.6 | 46.8 |
| | | (21%) | (44%) | (21%) | (19%) | (31%) | | |

*Table 5:* Cross-validation results for finite element mesh data.

## 5.5 Finite element mesh design

This application, first discussed by Dolšak and Muggleton (1992), concerns the division of an object into an appropriate number of regions for finite element simulation. Each edge in the object is cut into a number of intervals and the task is to learn to determine a suitable number – too fine a division requires excessive computation in the simulation, while too coarse a partitioning results in a poor approximation of the object's true behavior.

The data concern five objects with a total of 277 edges. The target relation mesh(A,B) specifies for each edge A the number of intervals B recommended by an expert, ranging from 1 to 12. Thirty background relations describe properties of each edge, such as its shape and its topological relationship to other edges in the object. Five trials are conducted, in each of which all information about one object is withheld, a definition learned from the edges in the remaining objects, and this definition tested on the edges in the omitted object.

Table 5 shows, for each trial, the number of edges on which the definitions learned by FOIL and FFOIL predict the number of intervals specified by the expert. Table 5 also shows published results on the mesh task for three other relational learning systems. The numbers of edges for which mFOIL and Golem predict the correct number of intervals are taken from (Lavrač and Džeroski, 1994). These are both general relational learning systems like FOIL, but FORS (Karalič, 1995), like FFOIL, is specialized for learning functional relations of this kind. Since the general relational learning systems could return multiple answers to the query mesh(e,X)? for edge e, only the first answer is used; this puts them at a disadvantage with respect to FOIL and FORS and accounts at least in part for their lower accuracy. Using a one-tailed sign test at the 5% level, FFOIL's accuracy is significantly higher than that achieved by FOIL and Golem, but no other differences are significant.

The time required by FFOIL for this domain is approximately three times that used by FOIL. This turnabout is caused by FFOIL's global pruning phase, which requires many literal eliminations in order to maximize overall accuracy on the training data. In one ply of the cross-validation, for instance, the initial definition, consisting of 30 clauses containing 64 body literals, fails to cover 146 of the 249 given tuples in the target relation mesh. After global pruning, however, the final definition has just 9 clauses with 15 body literals, and makes 101 errors on the training data.

## 6. Related Research

Mooney and Califf's (1995) recent system FOIDL has had a strong influence on the development of FFOIL. Three features that together distinguish FOIDL from earlier systems like FOIL are:

- Following the example of FOCL (Pazzani and Kibler, 1992), background relations are defined intensionally by programs rather than extensionally as tuple sets. This eliminates a problem in some applications for which a complete extensional definition of the background relations would be impossibly large.

- Examples of tuples that do not belong to the target relation are not needed. Instead, each argument of the target relation has a mode as above and FOIDL assumes *output completeness*, i.e., the tuples in the relation show all valid outputs for any inputs that appear.

- The learned definition is ordered and every clause ends with a cut.

Output completeness is a weaker restriction than functionality since there may be several correct answers to a standard query $R(c_1, c_2, ..., c_{n-1}, X)$?. However, the fact that each clause ends with a cut reduces this flexibility somewhat, since all answers to a query must be generated by a single clause.

Although FOIDL and FFOIL both learn ordered clauses with cuts, they do so in very different ways. FFOIL learns a clause, then a sequence of clauses to cover the remaining tuples, so that the first clause in the definition is the first clause learned. FOIDL instead follows Webb and Brkič (1993) in learning the *last* clause first, then prepending a sequence of clauses to filter out all exceptions to the learned clause. This strategy has the advantage that general rules can be learned first and still act as defaults to clauses that cover more specialized situations.

The principal differences between FOIDL and FFOIL are thus the use of intensional versus extensional background knowledge and the order in which clauses are learned. There are other subsidiary differences – for example, FOIDL never manipulates ⊖ bindings explicitly but estimates their number syntactically. However, in many ways FFOIL may be viewed as an intermediate system lying mid-way between FOIL and FOIDL.

FOIDL was motivated by the past tense task described in Section 5.4, and performs extremely well on it. The formulation of the task for FOIDL uses the relation past(A,B) to indicate that B is the past tense of verb A, together with the intensional background relation split(S,H,T) to denote all possible ways of dividing string S into substrings H and T. Definitions learned by FOIDL are compact and intelligible, and have a slightly higher accuracy (89.3%) than FFOIL's using the same ten sets of training and test examples. It will be interesting to see how the systems compare in other applications.

Bergadano and Gunetti (1993) first pointed out the advantages for learning systems of restricting relations to functions. Their FILP system assumes that all relations, both target and background, are functional, although they allow functions with multiple outputs. This assumption greatly reduces the number of literals considered when specializing a clause, leading to shorter learning times. (On the other hand, many of the tasks discussed in the previous section involve non-functional background relations and so would not satisfy FILP's functionality assumption.) In theory, FILP also requires an oracle to answer non-ground queries regarding unspecified tuples in the target and background relations, although this would not be required if all relevant tuples were provided initially. FILP guarantees that the learned definition is completely consistent with the given examples, and so is inappropriate for noisy domains such as those discussed in Sections 5.4 and 5.5.

In contrast to FFOIL and FOIDL, the definitions learned by FILP consist of unordered sets of clauses, despite the fact that the target relation is known to be functional. This prevents a clause from exploiting the context established by earlier clauses. In the gcd task (Section 5.3), a definition learned by FILP would require the bodies of both the second and third clauses to include a literal plus(...,...,...). In domains such as the past tense task, the complexity of definitions learned by FFOIL and FOIDL would be greatly increased if they were constrained to unordered clauses.

## 7. Conclusion

In this study, a mature relational learning system has been modified to customize it for functional relations. The fact that the specialized FFOIL performs so much better than the more general FOIL on relations of this kind lends support to Bergadano and Gunetti's (1993) thesis that functional relations are easier to learn. It is interesting to speculate that a similar improvement might well be obtainable by customizing other general first-order systems such as Progol (Muggleton, 1995) for learning functional relations.

Results from the quicksort experiments suggest that FFOIL scales better than general first-order systems when learning functional relations, and those from the past tense and mesh design experiments demonstrate its effectiveness in noisy domains.

Nevertheless, it is hoped to improve FFOIL in several ways. The system should be extended to multifunctions with more than one output variable, as permitted by both FILP and FOIDL. Secondly, many real-world tasks such as those of Sections 5.4 and 5.5 result in definitions in which the output variable is usually bound by being equated to a constant rather than by appearing in a body literal. In such applications, FFOIL is heavily biased towards constructing the next clause to cover the most frequent function value in the remaining tuples, as this binding tends to have the highest gain. By the time that the clause has been specialized to exclude exceptions, however, it can end up covering just a few tuples of the relation. If a few special cases could be filtered out first, clauses like this would be simpler and would cover more tuples of the target relation. A better learning strategy in these situations would seem to be to grow a new clause for every function value in the uncovered tuples, then retain the one with greatest coverage and discard the rest. This would involve an increase in computation but should lead to better, more concise definitions.

Although the conceptual changes in moving from FOIL to FFOIL are relatively slight, their effects at the code level are substantial (with only three of the 19 files that make up FOIL escaping modification). As a result it has been decided to preserve them as separate systems, rather than incorporating FFOIL as an option in FOIL. Both are available (for academic research purposes) by anonymous ftp from ftp.cs.su.oz.au, directory pub, file names foil6.sh and ffoil2.sh.

## Acknowledgements

This research was made possible by a grant from the Australian Research Council. Thanks to William Cohen, Ray Mooney, Michael Pazzani, and the anonymous reviewers for comments that helped to improve this paper.

## Appendix: Learned Definitions

The definition learned by FOIL appears on the left and that by FFOIL on the right. As the latter's default clauses are irrelevant for these tasks, they are omitted.

### List processing functions (Section 5.1)

(a) Using lists of length 3:

```
append([ ],B,B).
append(A,B,C) :- components(A,D,E),
        components(C,D,F), append(E,B,F).

last(A,B) :- components(A,B,[ ]).
last(A,B) :- components(A,C,D), last(D,B).

reverse(A,A) :- append(A,C,D),
        components(D,E,A).
reverse(A,B) :- last(A,C), last(B,D),
        components(A,D,E),
        components(B,C,F), reverse(E,G),
        del(D,B,G).

shift(A,B) :- components(A,C,D), del(C,B,D),
        append(D,E,B).

translate([ ],[ ]).
translate(A,B) :- components(A,C,D),
        components(B,E,F), translate(D,F),
        means(C,E).
```

```
append([ ],B,B) :- !.
append(A,B,C) :- components(A,D,E),
        append(E,B,F), components(C,D,F), !.

last(A,B) :- components(A,C,D), last(D,B), !.
last(A,B) :- member(B,A), !.

reverse(A,A) :- append(A,C,D),
        components(D,E,A), !.
reverse(A,B) :- components(A,C,D),
        reverse(D,E), append(F,D,A),
        append(E,F,B).

shift(A,B) :- components(A,C,D),
        append(E,D,A), append(D,E,B).

translate([ ],[ ]) :- !.
translate(A,B) :- components(A,C,D),
        translate(D,E), means(C,F),
        components(B,F,E).
```

(b) Using lists of length 4:

```
append([ ],B,B).
append(A,B,C) :- components(A,D,E),
        components(C,D,F), append(E,B,F).

last(A,B) :- components(A,B,[ ]).
last(A,B) :- components(A,C,D), last(D,B).

reverse(A,A) :- append(A,A,C), del(D,E,C).
reverse(A,B) :- components(A,C,D),
        reverse(D,E), append(F,D,A),
        append(E,F,B).

shift(A,B) :- components(A,C,D), del(C,B,D),
        append(D,E,B).
```

```
append([ ],B,B) :- !.
append(A,B,C) :- components(A,D,E),
        append(E,B,F), components(C,D,F), !.

last(A,B) :- components(A,C,D), last(D,B), !.
last(A,B) :- member(B,A), !.

reverse(A,A) :- append(A,C,D),
        components(D,E,A), !.
reverse(A,B) :- components(A,C,D),
        reverse(D,E), append(F,D,A),
        append(E,F,B).

shift(A,B) :- components(A,C,D),
        append(E,D,A), append(D,E,B).
```

```
translate([ ],[ ]).
translate(A,B) :- components(A,C,D),
        components(B,E,F), translate(D,F),
        means(C,E).
```

```
translate([ ],[ ]) :- !.
translate(A,B) :- components(A,C,D),
        translate(D,E), means(C,F),
        components(B,F,E).
```

## Quicksort and bubblesort (Section 5.2)

```
qsort([ ],[ ]).
qsort(A,B) :- components(A,C,D),
        partition(C,D,E,F), qsort(E,G),
        qsort(F,H), components(I,C,H),
        append(G,I,B).
```

```
qsort([ ],[ ]) :- !.
qsort(A,B) :- components(A,C,D),
        partition(C,D,E,F), qsort(E,G),
        qsort(F,H), components(I,C,H),
        append(G,I,B), !.
```

```
bsort([ ],[ ]).
bsort(A,A) :- components(A,C,[ ]).
bsort(A,B) :- components(A,C,D),
        components(B,C,E), bsort(D,E),
        components(E,F,G), lt(C,F).
bsort(A,B) :- components(A,C,D),
        components(B,E,F), bsort(D,G),
        components(G,E,H), lt(E,C),
        components(I,C,H), bsort(I,F).
```

```
bsort([ ],[ ]) :- !.
bsort(A,A) :- components(A,C,[ ]), !.
bsort(A,B) :- components(A,C,D), bsort(D,E),
        components(E,F,G),
        components(B,C,E), lt(C,F), !.
bsort(A,B) :- components(A,C,D), bsort(D,E),
        components(E,F,G),
        components(D,H,I),
        components(J,C,I), bsort(J,K),
        components(B,F,K), !.
bsort(A,B) :- components(A,C,D), bsort(D,E),
        components(F,C,E), bsort(F,B), !.
```

## Arithmetic functions (Section 5.3)

```
Ackermann(0,B,C) :- succ(B,C).
Ackermann(A,0,C) :- succ(D,A),
        Ackermann(D,1,C).
Ackermann(A,B,C) :- succ(D,A), succ(E,B),
        Ackermann(A,E,F),
        Ackermann(D,F,C).
```

```
Ackermann(0,B,C) :- succ(B,C), !.
Ackermann(A,0,C) :- succ(0,D), succ(E,A),
        Ackermann(E,D,C), !.
Ackermann(A,B,C) :- succ(D,A), succ(E,B),
        Ackermann(A,E,F),
        Ackermann(D,F,C), !.
```

```
gcd(A,A,A).
gcd(A,B,C) :- plus(B,D,A), gcd(B,A,C).
gcd(A,B,C) :- plus(A,D,B), gcd(A,D,C).
```

```
gcd(A,A,A) :- !.
gcd(A,B,C) :- plus(A,D,B), gcd(A,D,C), !.
gcd(A,B,C) :- gcd(B,A,C), !.
```